\documentclass[a4paper, 12pt, one column, aas_macros]{article}

\usepackage[english]{babel}
\usepackage[utf8]{inputenc}
\usepackage[T1]{fontenc}
\usepackage{aas_macros}

\usepackage[top=1.3cm, bottom=2.0cm, outer=2.5cm, inner=2.5cm, heightrounded,
marginparwidth=1.5cm, marginparsep=0.4cm, margin=2.5cm]{geometry}

\usepackage{graphicx} 
\usepackage[colorlinks=False]{hyperref} 
\usepackage{amsmath}  
\usepackage{amsfonts} %
\usepackage{amssymb}  %
\usepackage{mathptmx}
\usepackage{float}
\usepackage[square,numbers]{natbib}
\usepackage{caption}
\usepackage{lipsum}
\usepackage[ruled]{algorithm2e}
\title{\textbf{\Huge{A Survey On (Stochastic Fractal Search) Algorithm}}\vspace{-.8em}}
\author{\small{Mohammed ElKomy, Computer Engineering, mohammed.a.elkomy@f-eng.tanta.edu.eg}\vspace{1em}}
\date{}

\emergencystretch=\maxdimen
\usepackage[document]{ragged2e}
\usepackage{algpseudocode}

\DeclareUnicodeCharacter{0304}{\u{a}}
\begin{document}

    \maketitle

    \begin{abstract}
        \justifying
        \noindent
        Evolutionary Algorithms are {naturally} inspired {approximation} optimisation algorithms that usually interfere with science problems when common mathematical methods are unable to provide a good solution or finding the exact solution requires an {unreasonable} amount of time using traditional exhaustive search algorithms. The success of these population-based frameworks is mainly due to their flexibility and ease of adaptation to the most different and complex optimisation problems.\hspace{5pt}This paper presents a metaheuristic algorithm called
        Stochastic Fractal Search, inspired by the natural phenomenon of growth based on a
        mathematical concept called the {fractal}, which is shown to be able to explore the search space more efficiently.
        This paper also focuses on the algorithm steps and some example applications of engineering design optimisation problems commonly used in the literature being applied to the proposed algorithm.
    \end{abstract}

    \section*{Keywords}
    Optimisation, Exploration, Exploitation, Diffusion process, Fractal, Random fractals, Benchmark functions, Visual tracking, PID Controllers Tuning.

    \section*{Introduction}
    \justifying
    \noindent
    Throughout history problem-solving has been a major concern for science and industry, thousands of challenging problems have been proposed with varying levels complexity and granularity.
    Industrial and economic organizations are always looking for better solutions for the problems they encounter from, minimizing time, risk and cost, to maximizing profit and efficiency, which has been a persisting question.
    Also in engineering and scientific research there are certainly plenty of problems framed in an optimisation framework, including many variables acting under complex constraints such as maximum stress, maximum deflection,
    minimum load capacity, or geometrical configuration.
    Another major concern is the size of the search space increases dramatically while solving high-dimensional optimisation problems which hinders classical optimisation algorithms such as exhaustive search \citep{norvig} by consuming {unreasonable} amount of time due to the increase complexity.
    That's why metaheuristic and naturally inspired evolutionary algorithms are used as an optimisation approximation framework, while being able to produce robust solutions in reasonable amount of time without exhaustively searching the whole search space based on stochasticity.
    Those algorithms have satisfy two main important characteristics namely, intensification (or exploitation) and diversification (or exploration), Intensification represents the ability of the search algorithm to find the best candidates around the current best solutions, while diversification considers the efficiency of the algorithm in exploring the search space often using the randomization strategies.

    \noindent
    Recently, the scientific community has witnessed the cutting-edge breakthrough in computational intelligence including computer vision and natural language processing using numerical-based approaches, mostly driven by the advances in the integrated chips industry and graphical processing units (GPUs) for distributed computing.
    With this great development, metaheuristic algorithms, inspired by natural phenomena behaviors, received great attention, Among them, Genetic Algorithm (GA) \citep{GEN} based on the Darwin theorem of evolution, Particle Swarm Optimization (PSO) \citep{PSO} which mimics the flocks of birds searching for food and Ant colony (AC) \citep{ACO} is another optimisation algorithm inspired by the foraging behavior of ant colonies.
    \\
    This paper presents two algorithms \citep{SFS}, the first algorithm, each candidate solution simulate the branching property y of a dielectric breakdown, The second algorithm is the developed version of the first algorithm which tackles the disadvantages of the first algorithm.
    the second algorithm is divided into two processes called Diffusing and Updating processes, the first process is adapted from the first algorithm using the particle diffusion, while the second process is incorporated as an improvement to explore the search space more efficiently by introducing random perturbation to candidate solutions which in turn reduces the possibilty of being stuck in a local minima.

    \section*{Fractals}
\begin{figure}[H]
        \minipage{0.47\textwidth}
        \includegraphics[height=.6\linewidth]{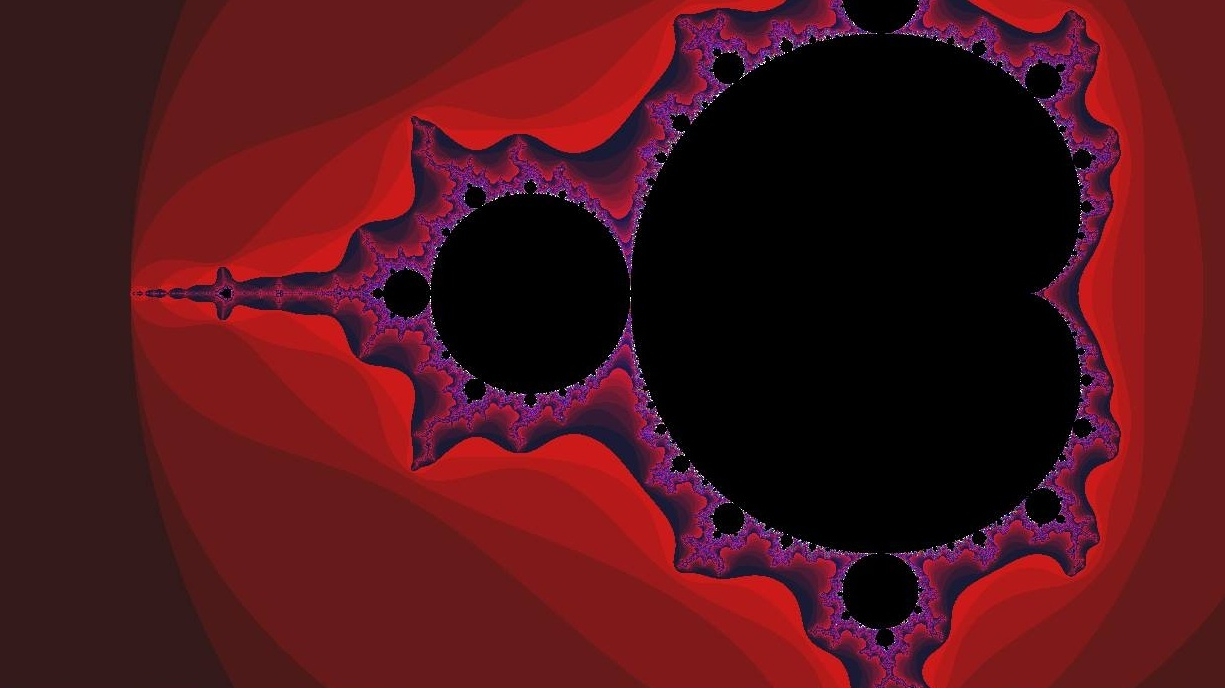}
        \vspace{-20pt}
        \caption{Mandelbrot Set}
        \label{mandel}
        \endminipage\hfill
        \minipage{0.47\textwidth}%
        \includegraphics[height=.6\linewidth,width=1\linewidth]{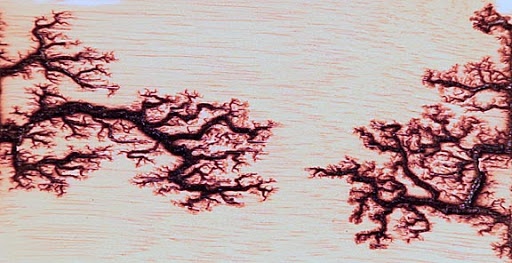}
        \vspace{-20pt}
        \caption{Dielectric Breakdown}
        \label{dielectric}
        \endminipage \hfill
    \end{figure}

    The property of an object or quantity which incorporates self-similarity on all scales\citep{SFS}, from the Latin word fr̄ctus which means 'broken' or 'fractured', Mathematically generated fractals are visually appealing recursive structures, for example Mandelbrot set  \hyperref[mandel]{(fig \ref{mandel})}  \citep{Mandelbrot} and Sierpinski triangle   \hyperref[triangle]{(fig \ref{triangle})} , There are some common methods to generate fractals, but in this work we are interested in generating random fractals as a core for the optimisation algorithm.
    \\
    Fractals are common in nature such as dielectric breakdown  \hyperref[dielectric]{(fig \ref{dielectric})} which is narrow discharge branchings due to lightning.

    \subsection*{Random Fractals}
    Random fractals are generated through modifying the iteration process according to a stochastic rule such as Levy flight, guassian walk or self-avoiding walks.
    The proposed search algorithm uses the Diffusion Limited Aggregation (DLA) method is the key operation used for the diffusion phase \hyperref[dla]{(fig \ref{dla})}  (\citep{SFS} fig 1).

    \begin{minipage}{0.45\linewidth}
        \begin{figure}[H]
            \center%
            \captionsetup{justification=centering}
            \includegraphics[height=.6\linewidth]{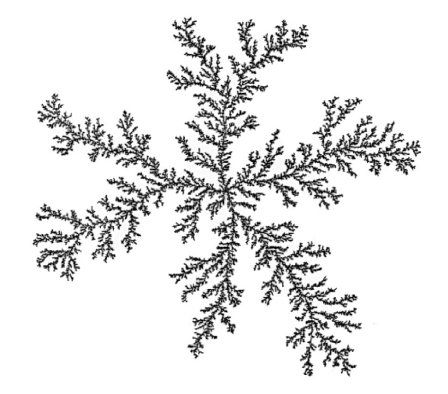}
            \vspace{-10pt}
            \caption{A simple fractal growth \\by DLA method.}
            \label{dla}
        \end{figure}
    \end{minipage}
    \begin{minipage}{0.45\linewidth}
        \begin{figure}[H]
            \center%

            \includegraphics[height=.6\linewidth]{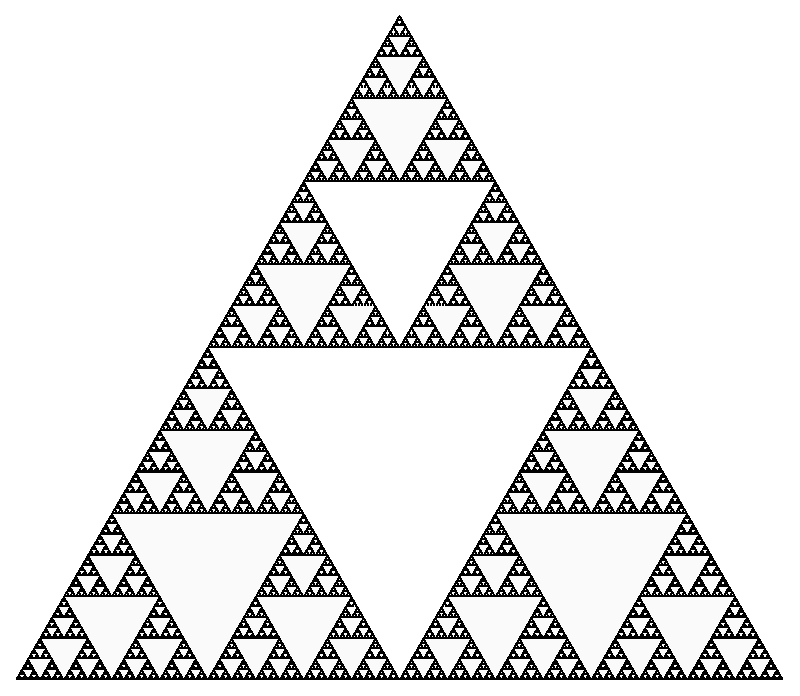}
            \vspace{-10pt}
            \caption{Sierpinski triangle}
            \label{triangle}
        \end{figure}
    \end{minipage}

    \section*{Fractal Search Algorithm}
    Based on our previous discussion, The first proposed optimisation algorithm incorporates both fractal growth through the DLA method and potential theory, particles are generated according to the following rules:
    \begin{enumerate}
        \item  Each particle has it's own electrical {potential} energy.
        \item  Each particle is subject to the diffusion process, generating other {random particles}, where the energy of the seed particle is distributed among the generated particles.
        \item  Only few of the {best particles} survive through successive generations, and the rest of the unsuccessful particles are dropped from the population.
    \end{enumerate}

    \begin{minipage}{0.5\linewidth}
        \noindent
        As detailed in (\citep{SFS} section 3), through generations each particle is diffused based on levy flight, A levy flight is a random-walk based on the levy distribution to model foraging in nature.
        For a particle with initial energy $E_i$, the diffusion process generates $p$ particles where the total potential energy of the generated particles is equal to the original seed particle potential energy. \hyperref[diffus]{(fig \ref{diffus})} (\citep{SFS} fig 2).
    \end{minipage}
    \begin{minipage}{0.4\linewidth}
        \begin{figure}[H]
            \center%
            \captionsetup{justification=centering}
            \includegraphics[width=1\linewidth]{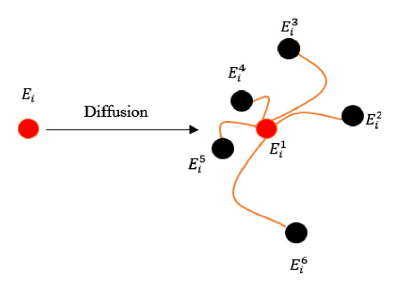}
            \vspace{-10pt}
            \caption{Diffusing a particle.}
            \label{diffus}
        \end{figure}
    \end{minipage}

    \vspace{10pt}
    \noindent The diffusion process switches randomly between the levy flight and gaussian distribution incorporating the advantages of both of them, where the levy flight for fast convergence while the gaussian distribution is used for improved exploitation.
    \\As shown in (\citep{SFS} section 4), Fractal search algorithm suffers from some disadvantages, mainly the number of tunable parameters and the lack of information exchange between particles, for that reason Stochastic Fractal Search (the second algorithm) addressing those drawbacks.
    \pagebreak
    \section*{Stochastic Fractal Search}
    As mentioned in the previous section, the disadvantages of Fractal search has been tackled in Stochastic Fractal Search(SFS) (\citep{SFS} section 4), introducing a phase called the updating process.

    \vspace{10pt}\noindent There are two main phases in SFS, namely, the diffusing process and the updating process.
    In the first process, as in Fractal Search, each particle diffuses around its current position to satisfy intensification (exploitation) of the search space, increasing the chance of finding the global minima, and also prevents being trapped in the local minima.
    In the second process, each particle updates its location in the search space based on the position of other particles, addressing the information exchange Fractal Search lacked, leading us to better diversification (exploration) of the search space.
    Another important modification is using a static diffusion process which means only the best generated particle from the diffusing process is considered, instead of the exponential growth of the population through generations.
    In addition to using a static diffusion process, the gaussian distribution is the only random walk considered rather than using both of the gaussian distribution and levy flight as in fractal search.
    Although levy flight proved fast convergence, gaussians are more promising in finding the global optimum.

    \begin{figure}[H]
        \center%
        \includegraphics[width=.9\textwidth]{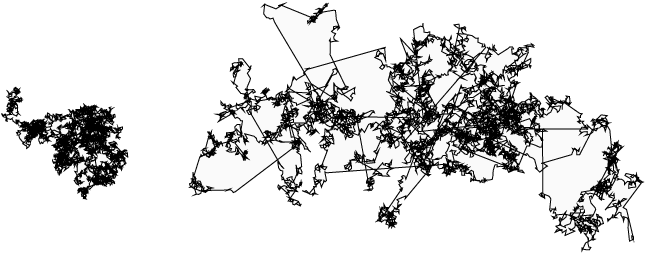}
        \caption{right:Levy flight, left:Gauss walk (\citep{flight})}
    \end{figure}

    \subsection*{Algorithm Steps And Pseudocode}

    \noindent Stochastic Fractal Search is fully presented in  \hyperref[alg]{(algorithm \ref{alg})}
    The first phase of the algorithm is the diffusion step which uses two different gaussian walks and switches between them randomly according to a tunable parameter given to the optimizer.
    \\\\The first gaussian walk, where $\mu_{BP}, BP$ is exactly Best Point found by the optimizer, $P_{i}$ the point from which the diffsion is done, $\varepsilon$ and $\varepsilon^{\prime}$ are uniformly distributed random numbers $[0,1]$.
    \begin{equation}
        G W_{1}=\operatorname{Gaussian}\left(\mu_{BP}, \sigma\right)+\left(\varepsilon \times BP-\varepsilon^{\prime} \times P_{i}\right)
        \label{11}
    \end{equation}
    The second gaussian walk, where $\mu_{\mathrm{P}} $ is point from which the diffusion is done, as in the equation below:
    \begin{equation}
        G W_{2}=\operatorname{Gaussian}\left(\mu_{\mathrm{P}}, \sigma\right)
        \label{12}
    \end{equation}
    For more improved localized search near the good candidates, by reducing the steps taken by the gaussian walks in (\ref{12}) and (\ref{11}), in the following equation:
    \begin{equation}
        \sigma=\left|\frac{\log (g)}{g} \times\left(P_{i}-B P\right)\right|
        \label{13}
    \end{equation}
    \\\\
    \noindent During the initialization, and for every newly created particle the boundaries of the search space have to be checked for a specific dimension $j$ for a particle $P$, where $UB$ is the upper bound vector, $LB$ is the lower bound vector, and $\varepsilon$  is a uniformly distributed random number [0,1], as in the following equation:
    \begin{equation}
        P_{j}=L B+\varepsilon \times(U B-L B)
        \label{14}
    \end{equation}

    \noindent The Linear ranking function to give probabilities to particles according to the fitness of each point, used twice in the   \hyperref[alg]{(algorithm \ref{alg})} , as in the following equation:
    \begin{equation}
        P a_{i}=\frac{\operatorname{rank}\left(P_{i}\right)}{N}
        \label{15}
    \end{equation}

    \noindent The first updating process uses the following equation, introducing elementwise updates For component $j$ and point $i$, $P_{i}^{\prime}$ is the newly updated point from the first updating step, as in the equation:
    \begin{equation}
        P_{i}^{\prime}(j)=P_{r}(j)-\varepsilon \times\left(P_{t}(j)-P_{i}(j)\right)
        \label{16}
    \end{equation}

    \noindent For the second updating process, for the point $P^{\prime}_{i}$ to be updated, and $P^{\prime}_{t}, P^{\prime}_{r}$ which are two randomly selected points, combining them together to perform a point vector replacement,\\ $\hat{\varepsilon}$ is a uniformly distributed random number [0,1] to switch between those two updating rules, $P_{i}^{\prime \prime}$ is the newly updated point from the second updating step, as described in the following equations:
    \begin{equation}
        P_{i}^{\prime \prime}=P_{i}^{\prime}-\hat{\varepsilon} \times\left(P_{t}^{\prime}-B P\right) \quad \mid \varepsilon^{\prime} \leqslant 0.5
        \label{17}
    \end{equation}
    \begin{equation}
        P_{i}^{\prime \prime}=P_{i}^{\prime}+\hat{\varepsilon} \times\left(P_{t}^{\prime}-P_{r}^{\prime}\right) \quad \mid \varepsilon^{\prime}>0.5
        \label{18}
    \end{equation}
    The pseudocode of the algorithm explains the steps and how each of those equations are used  \hyperref[alg]{(pseudocode \ref{alg})}.
    \section*{Recent SFS-Related Publications}
    \subsection*{PID controller design}
\begin{figure}[H]
        \center%
        \includegraphics[width=1\textwidth]{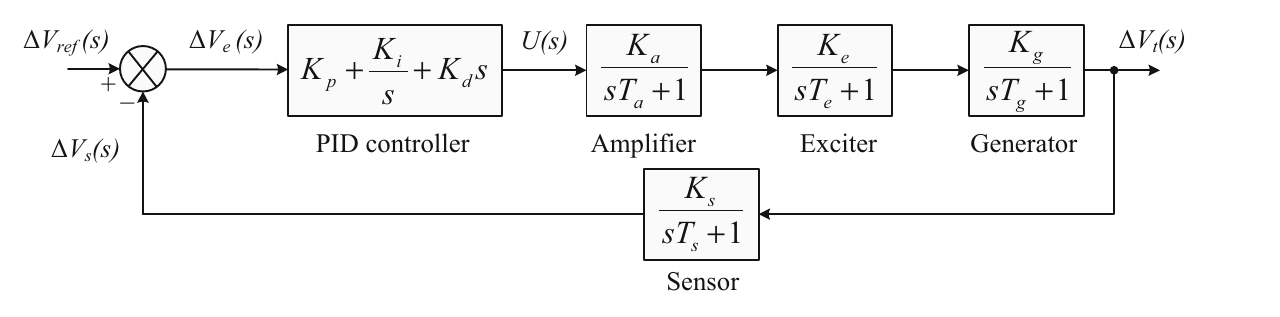}
        \caption{  Transfer function model of an AVR system with PID controller (\citep{PID}fig 2)}
        \label{PID2}
    \end{figure}

\begin{figure}[H]
        \center%
        \includegraphics[width=.8\textwidth]{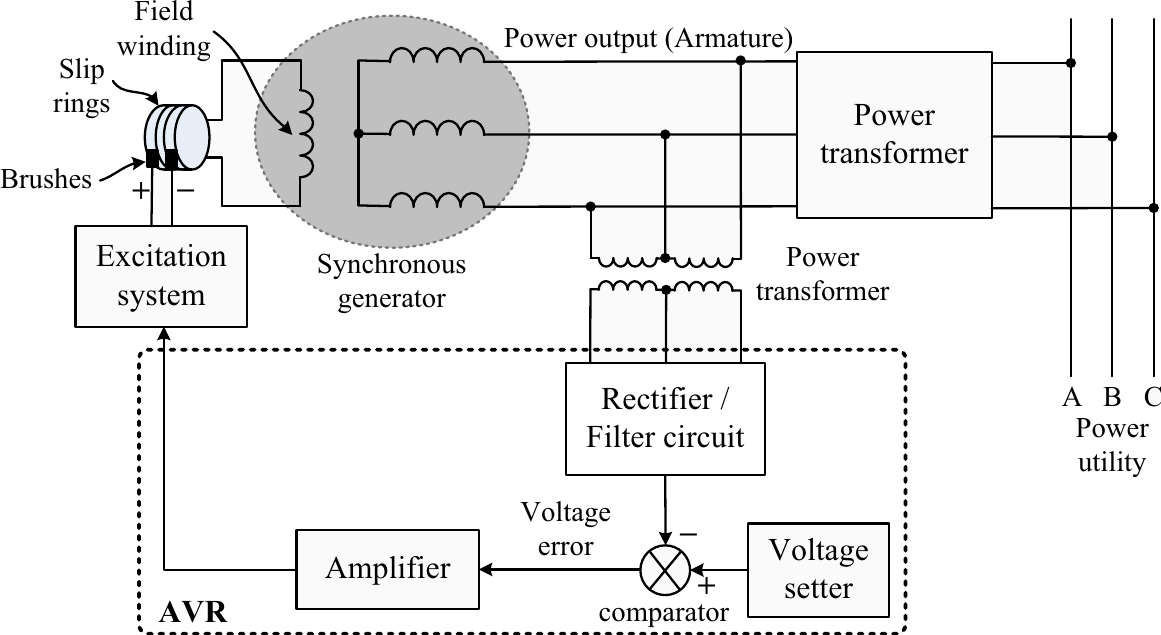}
        \caption{ Schematic diagram of an AVR system (\citep{PID} fig 1)}
        \label{PID1}
    \end{figure}

\begin{figure}[H]
        \center%
        \includegraphics[width=.95\textwidth]{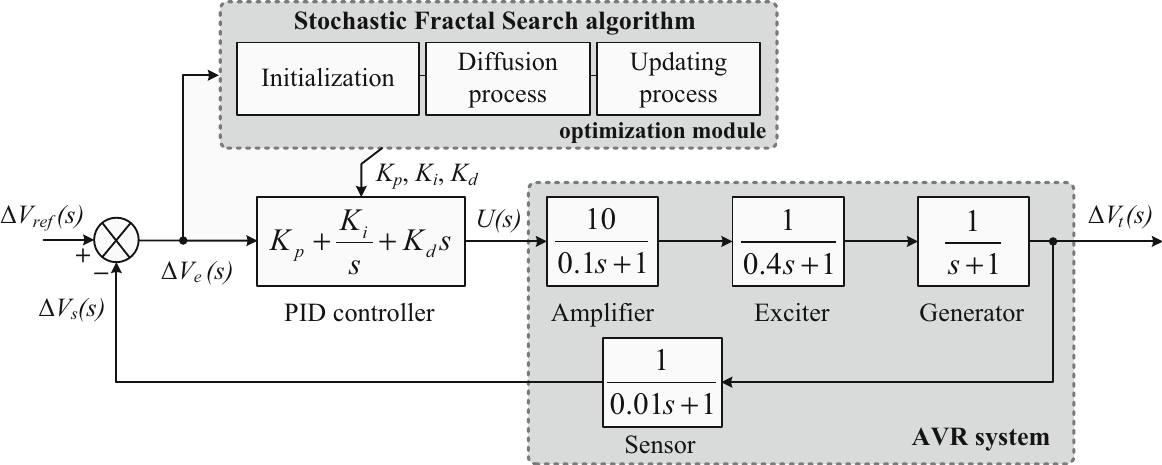}
        \caption{   Block diagram of an AVR system combined with SFS–PID controller(\citep{PID}fig 4)}
        \label{PID3}
    \end{figure}

    PID control algorithm is the most accepted approach for automatic voltage regulator systems \hyperref[PID2]{(fig \ref{PID2})}, {an automatic voltage regulator(AVR) } is equipment installed in power generation stations that sustains the output voltage at a desired voltage level under varying system conditions by controlling the excitation voltage of a synchronous generator \hyperref[PID1]{(fig \ref{PID1})}.
    despite that, it's still a challenging task for researchers to tune its parameters.
    \\A recent research \citep{PID} incorporated the {SFS} algorithm to tune the parameters of the {PID} controller  \hyperref[PID3]{(fig \ref{PID3})}, the motivation behind that was around $80\%$ of such controllers in service are suffering from {poorly} tuned controller gains.
    \\As in (\citep{PID} table 2), the SFS optimized PID controller renders a better dynamic response profile of the concerned power system than the existing alternative algorithms.

    \pagebreak

    \begin{algorithm}[H]
        \SetAlgoLined
        \BlankLine
        Initialize a population of N points\;
        \While{g < maximum generation or (stop criterion)}{


        \For{each Point $P_i$ in the system}
        {
        \textbf{Call} \textbf{Diffusion} Process with the following process:
        {

        q = (maximum considered number of diffusion).\\
        \For{ j = 1 to q}
        {
        \If{user applies the first Gaussian walk to solve the problem}
        {Create a new point based on Eq.(\ref{11}).}
            \If
            {user sets the second Gaussian Walks to solve the problem}
            {Create a new point based on Eq.(\ref{12}).}
                }

            }

        \vspace{10pt}
        \textbf{Call} \textbf{Updating} Process with the following process
        {

        \Begin
        {
        \vspace{5pt}\textbf{First Updating Process:}.\\

        First, all points are ranked based on Eq.( \ref{15}).

        \For{ each Point $P_i$ in the system }
        {

        \For{ each component $j$ in    $P_i$}
        {
        \If
        {rand[0,1] >  $Pa_i$}
        {Update the component inbased on Eq. (\ref{16}).
        }

            }

            }

            \vspace{5pt} \textbf{Second Updating Process:}.\\

            Once again, all points obtained by the first update process are ranked based on Eq. (\ref{15}).

            \For{ each Point $P'_i$ in the system }
            {

            \If
            {rand[0,1] >  $Pa'_i$}
            {Update the position based on Eqs. (\ref{17}) and (\ref{18}).}

                }
                }
        }

        }

            }

            \label{alg}

            \caption{Stochastic Fractal Search Pseudo Algorithm \citep[sec 4]{SFS}}
    \end{algorithm}

    \vspace{-5pt}

    \subsection*{Visual Object Tracking}
\begin{figure}[H]
        \center%
        \includegraphics[width=.85\textwidth]{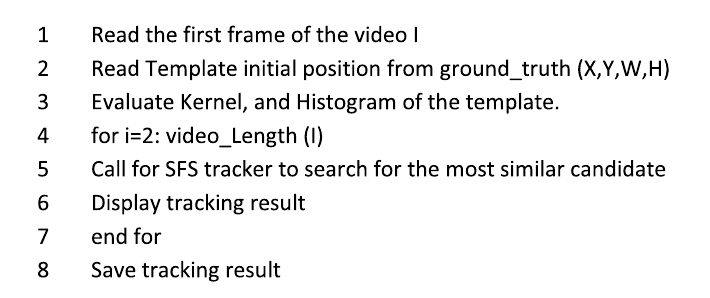}
        \caption{     Proposed SFS tracking algorithm  (\citep{Charef-Khodja2020}fig 3)}
        \label{CV_ALG}
    \end{figure}

\begin{figure}[H]
        \center%
        \includegraphics[width=.7\textwidth]{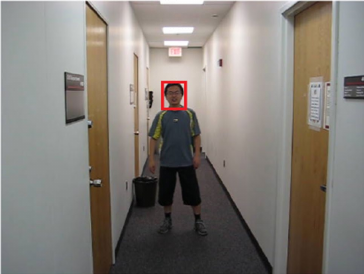}
        \caption{The Boy sequence suffers from scale variation, motion blur, fast motion, in- plane rotation, and out-of-plane rotation (\citep{Charef-Khodja2020}fig 4)}
        \label{boy}
    \end{figure}

    {Visual tracking} is a challenging computer vision problem because of many reasons such as, partial occlusion, fast motion, blur motion, object deformation, sudden feature changes, rotation, scale variation\dots etc, Example dataset \hyperref[boy]{(fig \ref{boy})}
    \\As in (\citep{Charef-Khodja2020}section 3) the target {localization} in every frame is considered as an {optimisation} problem.
    The pseudocode is shown in  \hyperref[CV_ALG]{(fig \ref{CV_ALG})},
    the object template is at first pinpointed inside a bounding box (BB), which is defined by a translation vector $V_i = (X_i, Y_i, W_i, H_i)$, where $(X_i , Y_i)$ denotes the cartesian coordinate of that BB, and $(W_i, H_i)$ are the fixed width and height of that BB defined from the ground truth of the first frame.
    Assuming that the target’s speed cannot exceed the target dimensions between adjacent frames, the reasonable size of the searching window is defined by the lower boundary $L_b$ and the upper boundary $U_b$.
    As depicted in (\citep{Charef-Khodja2020}section 3),

    \section*{Related work}
    Optimisation algorithms in the recent years have witnessed a dramatic increase of interest from scientists and engineers, because they provide acceptable solutions in a reasonable amount of time to many engineering and scientific problems.
    The development in optimisation algorithms focused on the evolutionary and metaheuristic algorithms in general because they are supported by natural phenomena which provably shows they actually survive the natural challenges.
    This dates back to the early stages of development, which incorporated the principles of evolution in problem-solving numerical techinques, from Genetic Algorithm (GA) \citep{GEN} based on the Darwin theorem of evolution, Ant colony (AC)\citep{ACO} mimicking the foraging behavior of ant colonies, Differential Evolution (DE) \citep{DE} is another heuristic based on simple yet powerful mathematical expressions, and particle Swarm Optimization (PSO) \citep{PSO} which simulates the flocks of birds searching for food.
    \newline This paper is an introduction to the algorithms presented in \citet{SFS}, namely, fractal search (\citep{SFS} section 3) and stochastic fractal search (\citep{SFS} section 4).
    Also mentioning applications in engineering and science using stochastic fractal search as an optimisation framework, Optimizing PID controller parameters for an automatic folder regulator \citep{Charef-Khodja2020},
    Another application \citep{PID} which used SFS as a localization backend for a visual tracking algorithm.

    \section*{Source Code}
    The source code for stochastic fractal search is already published by the original author on MATLAB Central File Exchange \href{https://www.mathworks.com/matlabcentral/fileexchange/47565-stochastic-fractal-search-sfs}{(here)}.
    In this paper we offer a python re-implementation \href{https://github.com/mohammed-elkomy/stochastic-fractal-search-python.git}{(here)} for SFS on github under the GPL-3.0 License, where almost all the well-known benchmark functions could be integrated in the pipeline.

    \section*{Conclusion}
    In this paper, we provided an introduction for fractal-based metaheuristic optimisation algorithms, which proved its ability to solve global optimisation problems qualitatively and quantitatively, the first algorithm incorporated the DLA method to generate new points based on the levy flight and gaussian distribution, switching between them randomly, on the other hand the latter algorithm employed a second phase for better convergence to the global optimum called the update step, which introduced random perturbation to candidate solutions.
    We also provided example applications that used SFS as an optimisation engine, outperforming alternative numerical-based optimisation. The first application addressed a control engineering problem, where SFS were used to optimise the parameters of a PID controller.
    The second application used SFS for object tracking, where in each frame the SFS is used to localize the location of the target object based on the previous knowledge from the previous frames.

    \bibliography{refs}
\end{document}